\documentclass[runningheads]{llncs}

\usepackage{eccv}

\usepackage{eccvabbrv}

\usepackage{graphicx}
\usepackage{booktabs}
\usepackage{multirow}
\usepackage{diagbox}
\usepackage{bbding}
\usepackage{makecell}

\newcommand{\tablestyle}[2]{\setlength{\tabcolsep}{#1}\renewcommand{\arraystretch}{#2}\centering}

\usepackage[dvipsnames]{xcolor}

\usepackage{pifont}
\newcommand{\cmark}{\ding{51}}%
\newcommand{\xmark}{\ding{55}}%

\usepackage[accsupp]{axessibility}  

\usepackage[pagebackref,breaklinks,colorlinks,citecolor=eccvblue]{hyperref}

\usepackage{orcidlink}

\begin{document}

\title{GhostNetV3: Exploring the Training Strategies for Compact Models}


\author{Zhenhua Liu\inst{1}$^\star$ \and
Zhiwei Hao\inst{2}\thanks{Equal contribution} \and
Kai Han\inst{1}$^{\star\star}$ \and
Yehui Tang\inst{1} \and
Yunhe Wang\inst{1}\thanks{Corresponding author}}

\authorrunning{Liu et al.}

\institute{Huawei Noah's Ark Lab, Beijing, China \\
\email{\{liu.zhenhua,kai.han,yehui.tang,yunhe.wang\}@huawei.com}\and
Beijing Institute of Technology, Beijing, China \\
\email{haozhw@bit.edu.cn}}

\maketitle

\begin{abstract}
  Compact neural networks are specially designed for applications on edge devices with faster inference speed yet modest performance.
  However, training strategies of compact models are borrowed from that of conventional models at present, which ignores their difference in model capacity and thus may impede the performance of compact models.
  In this paper, by systematically investigating the impact of different training ingredients, we introduce a strong training strategy for compact models. 
  We find that the appropriate designs of re-parameterization and knowledge distillation are crucial for training high-performance compact models, while some commonly used data augmentations for training conventional models, such as Mixup and CutMix, lead to worse performance.
  Our experiments on ImageNet-1K dataset demonstrate that our specialized training strategy for compact models is applicable to various architectures, including \text{GhostNetV3}, \text{MobileNetV2} and \text{ShuffleNetV2}.
  Specifically, equipped with our strategy, \text{GhostNetV3 1.3$\times$} achieves a top-1 accuracy of 79.1\% with only 269M FLOPs and a latency of 14.46ms on mobile devices, surpassing its ordinarily trained counterpart by a large margin.
  Moreover, our observation can also be extended to object detection scenarios. PyTorch code and checkpoints can be found at \href{https://github.com/huawei-noah/Efficient-AI-Backbones/tree/master/ghostnetv3_pytorch}{https://github.com/huawei-noah/Efficient-AI-Backbones}.
\end{abstract}

\section{Introduction}

To meet the limited memory and computational resource of edge devices (\eg, mobile phone), various efficient architectures~\cite{mobilenetv1, mobilenetv2, mobilenetv3, mobileone, mobilevit, ghostnet, ghostnetv2} have been developed.
For example, MobileNetV1~\cite{mobilenetv1} uses depth-wise separable convolutions to reduce computational cost.
MobileNetV2~\cite{mobilenetv2} introduces the residual connection, and MobileNetV3~\cite{mobilenetv3} further optimizes the architecture configurations via neural architecture search (NAS), which significantly improves performance of the model.
Another typical architecture is GhostNet~\cite{ghostnet}, which utilizes redundancy in feature and duplicates channels of the feature by using cheap operations.
Recently, GhostNetV2~\cite{ghostnetv2} further incorporates a hardware-friendly attention module to capture the dependence between long-range pixels and outperforms GhostNet by a significant margin.

Besides carefully designed model architectures, appropriate training strategies are also critical for remarkable performance.
For example, Wightman \etal~\cite{wightman2021resnet} improved the top-1 accuracy of ResNet-50~\cite{resnet} on ImageNet-1K~\cite{imagenet} from 76.1\% to 80.4\% by integrating advanced optimization and data augmentation approaches.
However, although considerable efforts have been made to explore more advance training strategies for conventional models (\eg, ResNet and Vision Transformer), little attention has been paid to compact models.
Since models with different capacities may have different learning preferences~\cite{tan2021efficientnetv2}, directly applying strategies designed for conventional models to train compact models is not appropriate.

\begin{figure}[t]
    \begin{center}
        \includegraphics[width=0.8\linewidth]{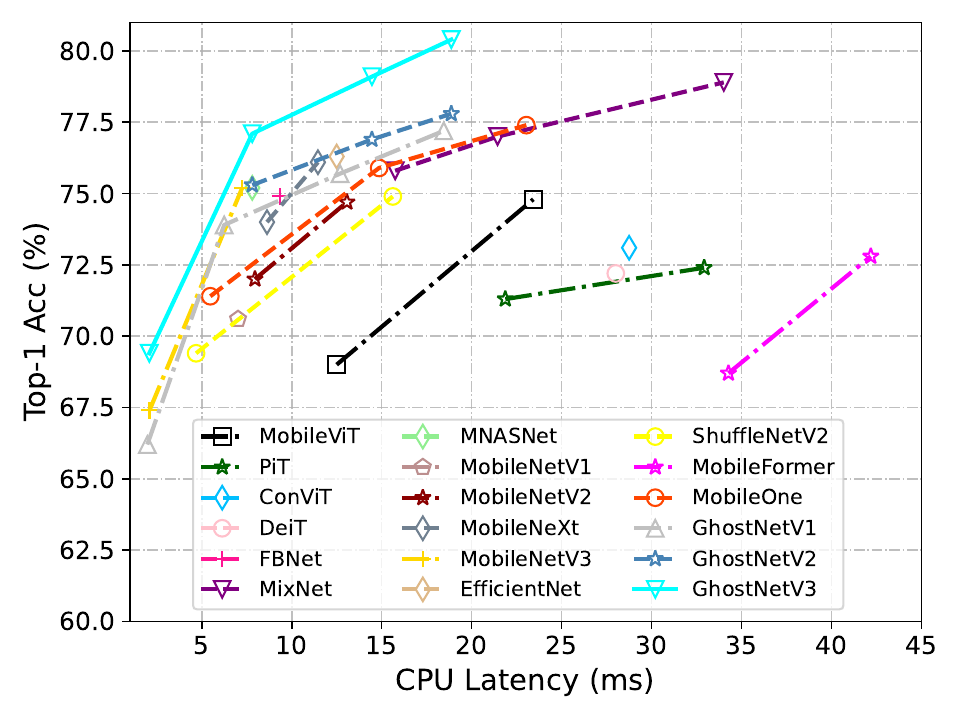}
    \end{center}
    \caption{The top-1 validation accuracy and the latency on CPU of various compact models on ImageNet dataset.}
    \label{fig:cpu}
\end{figure}

To bridge this gap, we systematically investigate several training strategies for compact models.
Specifically, our main attentions are paid on the key training settings as discussed in previous works~\cite{he2019bag,deit} including re-parameterization, knowledge distillation (KD), learning schedule and data augmentation.

\textbf{Re-parameterization.}
Depth-wise convolution and 1$\times$1 convolution are common components in compact model architectures due to their negligible memory and computational consumption.
Inspired by successful experiences in training conventional models~\cite{repvgg, ding2021diverse}, we employ the re-parameterization approach for these two compact modules to achieve better performance.
When training compact models, we introduce linear parallel branches into depth-wise convolution and 1$\times$1 convolution.
These additional parallel branches can be re-parameterized after training, bringing no extra cost at inference time.
To trade off the overall training cost against performance improvement, we compare the impact of different numbers of added branches.
Moreover, we find that a 1$\times$1 depth-wise convolution branch has a significant positive impact on the re-parameterization of 3$\times$3 depth-wise convolution.

\textbf{Knowledge distillation.}
It is challenging for compact models to achieve performance comparable to conventional models due to their limited model capacity.
Hence, KD~\cite{hinton2015distilling}, which employs a larger model as the teacher to guide the learning of compact models, is a proper approach to improve performance.
We empirically investigate the impact of several typical factors when training compact models using KD, such as the choice of teacher model and the setting of hyperparameters.
The results imply that an appropriate teacher model can significantly boost the performance of compact models.

\textbf{Learning schedule and data augmentation.}
We compare several training settings for compact models, including learning rate, weight decay, exponential moving average (EMA), and data augmentation.
Interestingly, not all the tricks designed for conventional models work well for compact models.
For instance, some widely-used data augmentation methods, such as \emph{Mixup} and \emph{CutMix}, actually detract from the performance of compact models.
We discuss their effect in Section~\ref{sec:exp} in detail.

Based on our investigation, we develop a specialized training recipe for compact models.
Experiments on ImageNet-1K dataset verify the superiority of our proposed recipe.
Specifically, the GhostNetV3 model trained with our recipe significant outperforms that trained with the previous strategy in terms of both top-1 accuracy and latency~(Figure~\ref{fig:cpu}).
Experiments on other efficient architectures such as MobileNetV2 and ShuffleNetV2 further confirm the generalizability of the proposed recipe.

The rest of the paper is organized as follows.
Section~\ref{sec:rel} reviews related works.
Section~\ref{sec:pre} introduces the architecture of GhostNetV2.
The training strategies are discussed in detail in Section~\ref{sec:rep}.
Then the plentiful experimental results are presented in Section~\ref{sec:exp}.
Section~\ref{sec:con} concludes the paper.

\section{Related works}
\label{sec:rel}
\subsection{Compact models}
It is challenging to design a compact model architecture with low inference latency and high performance simultaneously.
SqueezeNet~\cite{squeezenet} proposes three strategies to design a compact model, i.e., replacing 3$\times$3 filters with 1$\times$1 filers, decreasing the number of input channels to 3$\times$3 filters, and down-sampling late in the network to keep large feature maps.
These principles are constructive, especially the usage of 1$\times$1 convolution.
MobileNetV1~\cite{mobilenetv1} replaces almost all the filers with 1$\times$1 kernel and depth-wise separable convolutions, which dramatically reduces the computational cost.
MobileNetV2~\cite{mobilenetv2} further introduces the residual connection to the compact model, and constructs an inverted residual structure, where the intermediate layer of a block has more channels than its input and output.
To keep representation ability, a part of non-linear functions are removed.
MobileNeXt~\cite{mobilenext} rethinks the necessary of inverted bottleneck, and claims that the classic bottleneck structure can also achieve high performance.
Considering the 1$\times$1 convolution account for a substantial part of computational cost, ShuffleNet~\cite{shufflenet} replace it with group convolution.
The channel shuffle operation to help the information flowing across different groups.
By investigating the factors that affect the practical running speed, ShuffleNetV2~\cite{shufflenetv2} proposes a new hardware-friendly block.

MnasNet~\cite{mnasnet} and MobileNetV3~\cite{mobilenetv3} search the architecture parameters, such as model width, model depth, convolutional filter's size, \etc.
By leveraging the feature's redundancy, GhostNet~\cite{ghostnet} replaces half channels in 1$\times$1 convolution with cheap operations.
GhostNetV2~\cite{ghostnetv2} proposes a dubbed DFC attention based on fully-connected layers, which can not only execute fast on common hardware but also capture the dependence between long-range pixels.
Until now, the series of GhostNets are still the SOTA compact models with a good trade-off between accuracy and speed.

Since ViT~\cite{vit} (DeiT) has made a great success on computer vision tasks, researchers have made efforts to design compact transformer architecture for mobile devices.
MobileFormer~\cite{mobileformer} proposes a compact cross attention to model the two-way bridge between MobileNet and transformer.
MobileViT~\cite{mobilevit} takes the successful experiences in compact CNNs and replaces local processing in convolutions with global processing using transformers.
However, the transformer-based compact models suffer from high inference latency on mobile devices due to the complex attention operation.

\begin{figure}[ht]
    \centering
    \begin{subfigure}{0.305\linewidth}
        \includegraphics[width=\textwidth]{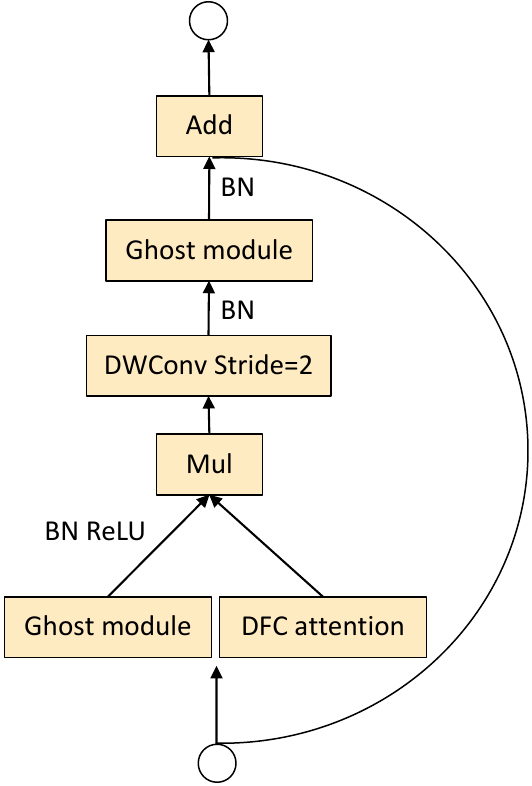}
        \caption{GhostNetV2}
        \label{fig:ghostnetv2}
    \end{subfigure}
    \hspace{0.01\linewidth}
    \begin{subfigure}{0.61\linewidth}
        \includegraphics[width=\textwidth]{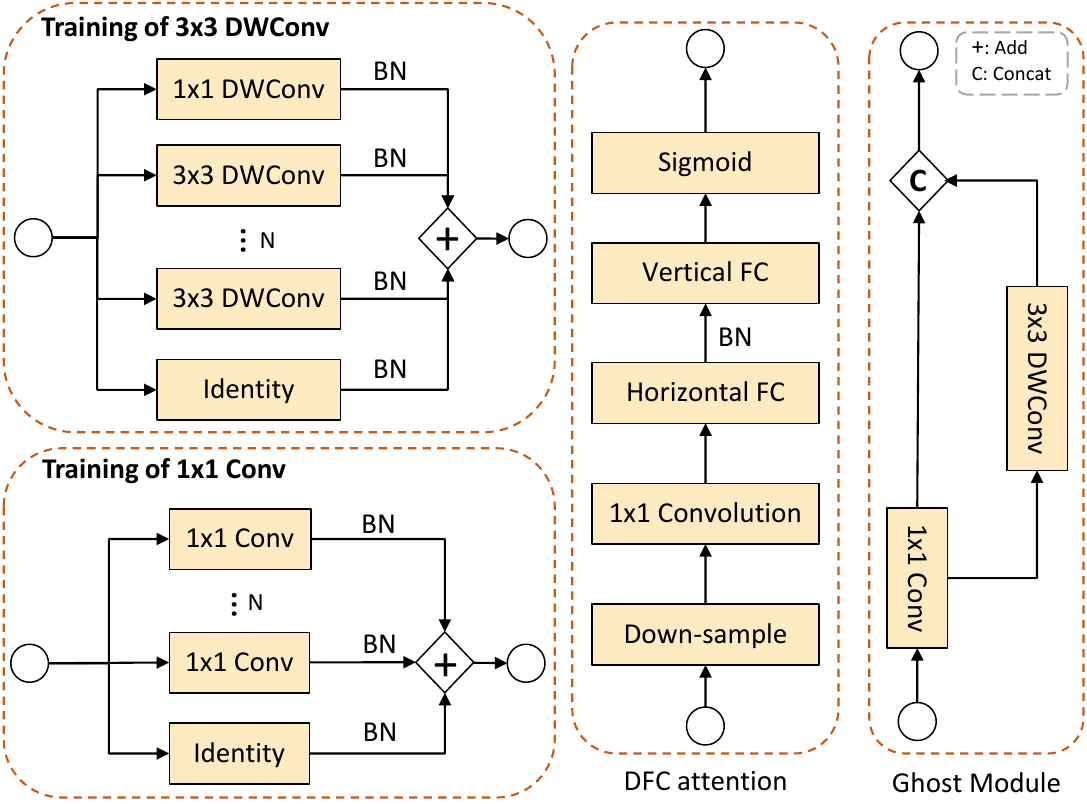}
        \caption{GhostNetV3}
        \label{fig:repara}
    \end{subfigure}
    \caption{The architectures of GhostNetV2 and GhostNetV3.}
\end{figure}

\subsection{Bag of tricks for training CNNs}
There are some works that focusing on improving training strategies to improve the performance of various models.
He~\etal~\cite{he2019bag} discuss several tricks that are useful for efficient training on hardware and propose a new model architecture tweaks for ResNet.
Wrightman~\etal~\cite{wightman2021resnet} re-evaluate the performance of the vanilla ResNet-50 when trained with novel optimization and data-augmentation methods.
They share the competitive training settings and pre-trained model in the timm open-source library.
With their training recipe, a vanilla ResNet-50 model achieves 80.4\% top-1 accuracy.
Chen~\etal~\cite{chen2021empirical} investigate the effects of several fundamental components for training self-supervised ViT.
However, all these attempts are designed for large models or self-supervised models.
Directly transferring them to compact models is inappropriate because of their different model capacity~\cite{tan2021efficientnetv2}.

\section{Preliminary}
\label{sec:pre}
GhostNets (GhostNetV1 and GhostNetV2) are the state-of-the-art compact models designed for efficient inference on mobile devices.
Their key architecture is the Ghost module, which can replace the original convolution by generating more feature maps from cheap operations.

In the ordinary convolution, the output feature $Y$ is obtained by $Y=X\ast W$, where $W\in \mathbb{R}^{c_\text{out} \times c_\text{in} \times k \times k}$ is the convolution kernel and $X$ is the input feature.
$c_{in}$ and $c_{out}$ denote the input and output channel dimension, respectively.
$k$ is the kernel size and $\ast$ denotes convolution operation.
A Ghost module reduces the number of parameters and computational cost of ordinary convolution in two steps.
It first produces \emph{intrinsic} features $Y'$, whose channel dimension is smaller than the original feature $Y$.
Then the cheap operation (\eg, depth-wise convolution) is applied on the intrinsic features $Y'$ to generate \emph{ghost} features $Y''$.
The final output is obtained by concatenating the \emph{intrinsic} and \emph{ghost} features along the channel dimension, which can be formulated as:
\begin{align}
     & Y'=X \ast W_p,             \\
     & Y={\rm Cat}(Y',X\ast W_c),
\end{align}
where $W_p$ and $W_c$ denote the parameters in primary convolution and cheap operations, respectively.
``Cat'' denotes the concatenating operation.
A whole GhostNet model is constructed by stacking multiple Ghost modules.

GhostNetV2 enhances compact models by designing an efficient attention module, \ie, DFC attention.
Considering that compact models such as GhostNet usually use small-kernel convolutions, \eg, 1$\times$1 and 3$\times$3, they have weak ability to extract global information from input features.
GhostNetV2 employs a simple fully-connected layer to capture the long-range spatial information and generate an attention map.
For computational efficiency, it decouples the global information into horizontal and vertical directions and aggregates pixels along the two directions, respectively.
As shown in Figure~\ref{fig:ghostnetv2}, by equipping Ghost module with DFC attention, GhostNetV2 can extract global and local information effectively while achieves a better trade-off between accuracy and computational complexity.

\section{Training strategies}
\label{sec:rep}
Our goal is to exploring the training strategies without changing the inference network architectures to keep the small model size and fast speed of compact models. We empirically investigate the key factors for training neural networks including learning schedule, data augmentation, re-parameterization and knowledge distillation.

\subsection{Re-parameterization}
Re-parameterization has proved its effectiveness in conventional convolutional models~\cite{repvgg, ding2021diverse}.
Inspired by their success, we introduce re-parameterization into compact models by adding repetitive branches equipped with BatchNorm layers.
Our design of re-parameterization GhostNetV3 is presented in Figure~\ref{fig:repara}.
It is worth noting that we introduce a 1$\times$1 depth-wise convolution branch into re-parameterized 3$\times$3 depth-wise convolution.
Experimental results confirm its positive effect on the performance of compact models.
Furthermore, the experiments thoroughly explore the optimal number of repetitive branches.

At inference time, the repetitive branches can be removed via an inverse re-parameterization process.
Since convolution and BatchNorm operations are both linear during inference, they can be folded into a single convolution layer, whose weight matrix is denoted as $\widehat{\mathbf{W}}\in \mathbb{R}^{c_\text{out}\times c_\text{in} \times k \times k}$ and bias is denoted as $\widehat{\mathbf{b}}\in \mathbb{R}^{c_\text{out}}$.
After that, folded weights and biases in all branches can be re-parameterized into $\mathbf{W}_{\text{rep}}=\sum_{i}\widehat{\mathbf{W}_i}$ and bias $\mathbf{b}_{\text{rep}}=\sum_{i}\widehat{\mathbf{b}_i}$, respectively, where $i$ is the index of repetitive branches.

\subsection{Knowledge distillation}

KD is a widely used model compression method, where the predictions of a large pre-trained teacher model is regarded as the learning target of a tiny student model.
Given a sample $x$ with label $y$, representing the corresponding logits predicted by the student and the teacher model using $\Gamma_\text{s}(x)$ and $\Gamma_\text{t}(x)$, respectively, the total loss function of KD can be formulated as:
\begin{equation}
    \mathcal{L}_{\text{total}}=(1-\alpha)\mathcal{L}_{\text{ce}}(\Gamma_\text{s}(x),y)+\alpha\mathcal{L}_{\text{kd}}(\Gamma_\text{s}(x), \Gamma_\text{t}(x)),
\end{equation}
where $\mathcal{L}_{\text{ce}}$ and $\mathcal{L}_{\text{kd}}$ denote the cross-entropy loss and KD loss respectively.
$\alpha$ is a balancing hyperparameter.

Usually, the Kullback-Leibler divergence function is adopted as the KD loss, which can be represented as:
\begin{equation}
    \mathcal{L}_{\text{kd}}=\tau^2 \cdot \text{KL}(\text{softmax}(\Gamma_\text{s}(x))/\tau, \text{softmax}(\Gamma_\text{t}(x))/\tau),
\end{equation}
where $\tau$ is a label smoothing hyperparameter termed temperature.
In our experiments, we study the impact of different settings of hyperparameters $\alpha$ and $\tau$ on the performance of compact models.

\subsection{Learning schedule}

Learning rate is a critical parameter in the optimization of neural networks.
There are two commonly used learning rate schedules: \emph{step} and \emph{cosine}.
The \emph{step} schedule decreases the learning rate linearly, whereas \emph{cosine} schedule reduces the learning rate slowly at the beginning, becomes almost linear in the middle, and slows down again at the end.
This work extensively investigates the impact of both learning rate and learning rate schedule on compact models.

Exponential moving average (EMA) has recently emerged as an effective approach to improve the validation accuracy and increase the robustness of models.
Specifically, it gradually averages parameters of a model during the training time.
Suppose parameters of the model at step $t$ is $\mathbf{W}_t$, the EMA of the model is computed as:
\begin{equation}
    \overline{\mathbf{W}}_t=\beta\cdot \overline{\mathbf{W}}_{t-1}+(1-\beta)\cdot \mathbf{W}_t,
\end{equation}
where $\overline{\mathbf{W}}_t$ represents the parameters of the EMA model at step $t$ and $\beta$ is a hyperparameter.
We study the effect of EMA in Section~\ref{sec:ema}.

\subsection{Data augmentation}

Various data augmentation approaches have been proposed to promote performance of conventional models.
Among them, \emph{AutoAug} scheme~\cite{autoaugment} adopts 25 combinations of sub-strategy, each of which contains two transformations.
For each input image, a sub-strategy combination is randomly selected, and the decision of whether to apply each transformation in the sub-strategy is determined by a certain probability.
\emph{RandomAug} method~\cite{randaugment} proposes a random augmentation approach where all sub-strategies are selected with the same probability.
Image aliasing methods like \emph{Mixup}~\cite{mixup} and \emph{CutMix}~\cite{cutmix} fuse two images to generate a new image.
Specifically, \emph{Mixup} trains a neural network on convex combinations of pairs of examples and their labels, whereas \emph{CutMix} randomly removes a region from one image and replace the corresponding area with a patch from another image.
\emph{RandomErasing}~\cite{randomerasing} randomly selects a rectangle region in an image and replaces its pixels with random values.

In this paper, we evaluate various combinations of the above data augmentation approaches and find that some commonly used data augmentation methods for training conventional models, such as \emph{Mixup} and \emph{CutMix}, are not appropriate for training compact models.

\section{Experimental results}
\label{sec:exp}
In our basic training strategy, we use a mini-batch size of 2048 and employ LAMB~\cite{lamb} for model optimization over 600 epochs. The initial learning rate is 0.005, and the \emph{cosine} learning schedule is adopted. Weight decay and momentum are set to 0.05 and 0.9, respectively. We use a decay factor of 0.9999 for the exponential moving average (EMA), where random augmentation and random erasing are applied for data augmentation. In this section, we explore these training strategies and unveil insights for training compact models. All experiments are conducted on the ImageNet dataset~\cite{imagenet} using 8 NVIDIA Tesla V100 GPUs.

\subsection{Re-parameterization}

To better comprehend the advantages of integrating re-parameterization into the training of compact models, we perform an ablation study to assess the impact of re-parameterization on different sizes of GhostNetV3. The results are presented in Table~\ref{tab:wrep}. The adoption of re-parameterization, while keeping other training settings unchanged, results in a significant improvement in performance compared to directly training the original GhostNetV3 models.

\begin{table}[!t]
  \begin{minipage}[t]{0.37\linewidth}
    \centering
    \caption{Top-1 accuracy of different versions of GhostNetV3 models trained with or without re-parameterization.}
		\begin{tabular}{ccccc}
			\toprule
			Model size & 1.0$\times$ & 1.3$\times$ & 1.6$\times$ \\
			\midrule
			w/o Rep    & 75.3        & 76.9        & 77.8      \\
			w/ ReP     & 76.1        & 77.6        & 78.7      \\
			\bottomrule
		\end{tabular}
		\label{tab:wrep}
  \end{minipage}
  \hfill
  \begin{minipage}[t]{0.57\linewidth}
    \centering
    \caption{Comparison of top-1 accuracy for different values of re-parameterization factor $N$. 'DW' denotes depth-wise convolution.}
    \begin{tabular}{cccc}
        \toprule
        Number of branches & w/o 1$\times$1DW & w/ 1$\times$1DW \\
        \midrule
        N=1                & 77.3      & 77.0     \\
        N=2                & 77.2      & 77.3     \\
        N=3                & 77.1      & 77.6    \\
        N=4                & 77.0      & 77.5     \\
        N=5                & 77.0      & 77.6     \\
        \bottomrule
    \end{tabular}
    \label{tab:rep}
  \end{minipage}
\end{table}

Additionally, we compare different configurations of the re-parameterization factor $N$ and the results are presented in Table~\ref{tab:rep}. As indicated in the results, the 1$\times$1 depth-wise convolution plays a crucial role in re-parameterization. If the $1\times$1 depth-wise convolution is not used in the re-parameterized model, its performance even decreases as the number of branches increases. In contrast, when equipped with a 1$\times$1 depth-wise convolution, the GhostNetV3 model achieves a peak top-1 accuracy of 77.6\% when $N$ is 3, and further increasing the value of $N$ brings no additional improvement in performance. Therefore, the re-parameterization factor $N$ is set to 3 in subsequent experiments to achieve better performance.

\subsection{Knowledge distillation}

In this section, we assess the impact of knowledge distillation on the performance of GhostNetV3. 
Specifically, ResNet-101~\cite{resnet}, DeiT-B~\cite{deit}, and BeiTV2-B~\cite{peng2022beit} are adopted as the teacher, achieving top-1 accuracies of 77.4\%, 81.8\%, and 86.5\%, respectively. 
The results in Table~\ref{tab:teacher} highlight performance variations with different teacher models. 
Notably, superior teacher performance correlates with improved GhostNetV3 performance, underscoring the importance of a well-performing teacher model in knowledge distillation with compact models.

\begin{table}[!t]
  \hspace{0.01\linewidth}
  \begin{minipage}[t]{0.5\linewidth}
    \centering
    \caption{Comparison of Top-1 accuracy for various teachers and $\alpha$ in knowledge distillation.}
    \begin{tabular}{cccc}
        \toprule
        \diagbox{Teacher}{$\alpha$}   & 0.5    & 0.9   & 1.0   \\
        \midrule
        ResNet-101~\cite{resnet}      & 78.46  & 78.53 & 78.20 \\
        DeiT-B~\cite{deit}            & 78.61  & 78.55 & 78.32 \\
        BEiTV2-B~\cite{peng2022beit}  & 79.13  & 79.02 & 78.86 \\
        \bottomrule
    \end{tabular}
    \label{tab:teacher}
  \end{minipage}
  \hfill
  \begin{minipage}[t]{0.4\linewidth}
    \centering
    \caption{Comparison of Top-1 accuracy for different $\alpha$ and temperature $\tau$ in knowledge distillation.}
    \begin{tabular}{ccccc}
        \toprule
        \diagbox{$\alpha$}{$\tau$} & 1           & 2     & 4     \\
        \midrule
        0.5                        & 79.13 & 77.98 & 77.91 \\
        1.0                        & 78.94       & 77.92 & 77.70 \\
        \bottomrule
    \end{tabular}
    \label{tab:kd1}
  \end{minipage}
  \hspace{0.01\linewidth}
\end{table}

We additionally compare different settings of hyperparameters in the KD loss with BEiTV2-B as the teacher.
The results in Table~\ref{tab:kd1} suggest that a low temperature value is preferable for compact models.
Furthermore, it is worth noting that when the KD loss is used alone (\ie $\alpha$=1.0), there is a noticeable decline in the top-1 accuracy.

We also explore the impact of combining re-parameterization and knowledge distillation on the performance of GhostNetV3. As shown in Table~\ref{tab:repkd}, the results indicate a significant improvement in performance (up to 79.13\%) due to the utilization of knowledge distillation. Moreover, it emphasizes the importance of the 1$\times$1 depth-wise convolution in re-parameterization. These findings underscore the importance of investigating various techniques and their potential combinations to enhance the performance of compact models.

\begin{table}[!t]
  \begin{minipage}[t]{0.48\linewidth}
    \centering
    \caption{Comparison of Top-1 accuracy for combining re-parameterization and knowledge distillation.}
    \begin{tabular}{cccc}
        \toprule
        Number of branches & 1           & 2           & 3           \\
        \midrule
        w/o 1$\times$1DW   & 78.59 & 78.73 & 78.85 \\
        w/ 1$\times$1DW    & 78.80 & 79.04 & 79.13 \\
        \bottomrule
    \end{tabular}
    \label{tab:repkd}
  \end{minipage}
  \hfill
  \begin{minipage}[t]{0.5\linewidth}
    \centering
    \caption{Top-1 accuracy of different learning rate schedule for GhostNetV2. The number after '\emph{cosine}' denotes the minimum of the learning rate.}
    \begin{tabular}{cccc}
        \toprule
        lr schedule & \emph{step} & \emph{cosine}\_0 & \emph{cosine}\_1e-5 \\
        \midrule
        w ReP\&KD   & 78.40       & 79.13            &     78.97                \\
        \bottomrule
    \end{tabular}
    \label{tab:lr_sched}
  \end{minipage}
\end{table}

\subsection{Learning schedule}
\begin{figure}[ht]
    \begin{center}
        \includegraphics[width=0.7\linewidth]{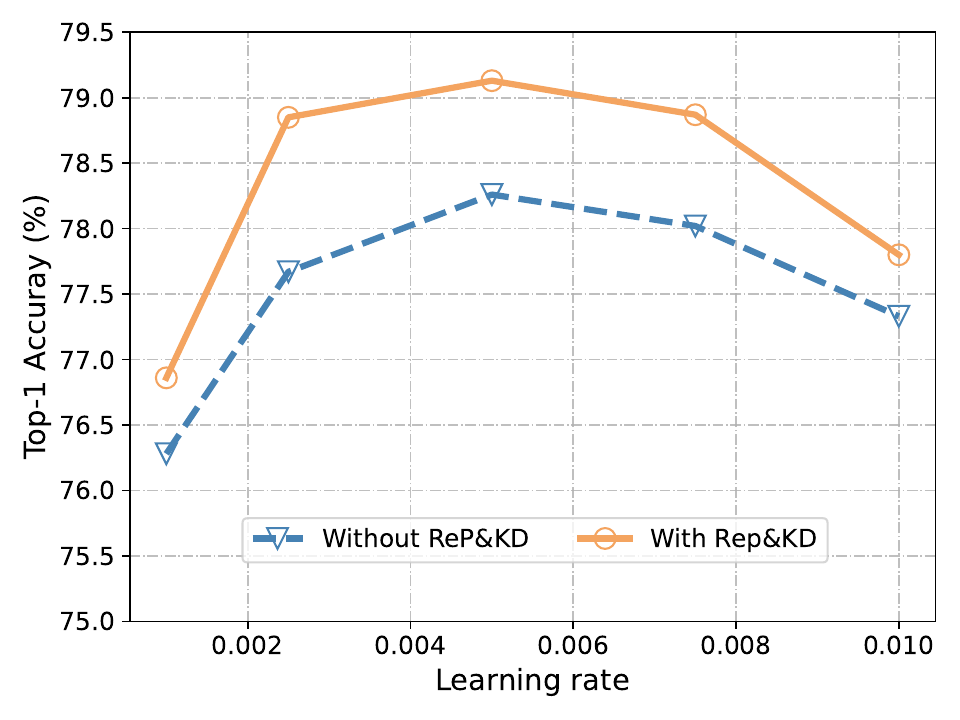}
    \end{center}
    \caption{The top-1 validation accuracy for various learning rates of GhostNetV3.}
    \label{fig:lr}
\end{figure}

\paragraph{Learning rate schedule.}
Figure~\ref{fig:lr} illustrates the experimental results adopting different scheduling schemes of the learning rate, both with and without re-parameterization and knowledge distillation. It is observed that both small and large learning rates have a detrimental effect on performance. Therefore, a learning rate of 5e-3 is chosen for the final experiments.

The \emph{step} and \emph{cosine} learning rate schedules are compared in Table~\ref{tab:lr_sched}.
It is observed that the \emph{cosine} learning rate schedule achieves the highest top-1 accuracy.
This underlines the effectiveness of a well-designed \emph{cosine} learning rate schedule in promoting the performance of compact models.

\paragraph{Weight decay.}

The effect of weight decay on the top-1 accuracy of GhostNetV3 is shown in Table~\ref{tab:wd}.
The results indicate that a large weight decay significantly diminishes the model's performance.
Therefore, we have retained a weight decay value of 0.05 for GhostNetV3, given its effectiveness for compact models.

\begin{table}[ht]
    \tablestyle{4pt}{1.0}
    \centering
    \caption{Top-1 accuracy achieved with different weight decay settings on the ImageNet dataset.}
    \begin{tabular}{cccccc}
        \toprule
        Weight decay & 0.01        & 0.03        & 0.05        & 0.07  & 0.1      \\
        \midrule
        w ReP\&KD    & 78.63 & 78.85 & 79.13 & 79.01  & 78.83 \\
        \bottomrule
    \end{tabular}
    \label{tab:wd}
\end{table}

\begin{figure}[ht]
    \begin{center}
        \includegraphics[width=0.7\linewidth]{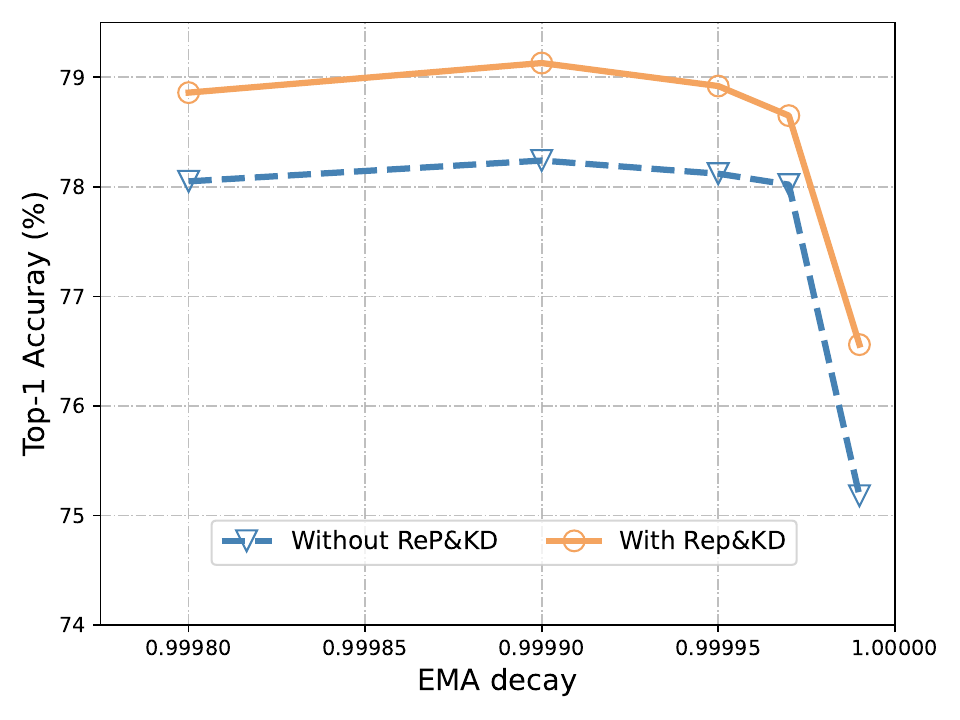}
    \end{center}
    \caption{The top-1 accuracy achieved with various decay values of EMA.}
    \label{fig:ema}
\end{figure}

\paragraph{EMA.}
\label{sec:ema}
In Figure~\ref{fig:ema}, it can be observed that when the decay of EMA is 0.99999, there is a decline in performance regardless of whether re-parameterization and knowledge distillation techniques are used or not. We speculate that this is due to the weakening effect of the current iteration when the decay value is too large. For compact models, a decay value of 0.9999 or 0.99995 is deemed appropriate, which is similar to that of a conventional model.

\subsection{Data Augmentation}

To compare the impact of different data augmentation schemes on the performance of lightweight models, we train cnn-based GhostNetV3 and ViT-based DeiT-tiny models under distinct augmentation strategies.
The results are presented in Table~\ref{tab:aug}.
It is observed that \emph{random augmentation} and \emph{random erasing} are advantageous for both GhostNetV3 and DeiT-tiny.
Conversely, \emph{Mixup} and \emph{CutMix} have a detrimental effect, which are then considered unsuitable for compact models.

\begin{table}[ht]
    \centering
    \tablestyle{5pt}{1.0}
    \caption{comparison results of different combinations of data augmentation schemes on GhostNetV3 and DeiT-tiny.}
    \begin{tabular}{ccccccc}
        \toprule
        \emph{AutoAug} & \emph{RandAug} & \emph{Mixup} & \emph{CutMix} & \emph{Erasing} & GhostNetV3 & DeiT-tiny  \\
        \midrule
        \cmark         & \xmark         & \xmark       & \xmark        & \xmark         & 78.78 &76.33          \\
        \xmark         & \cmark         & \xmark       & \xmark        & \xmark         & 79.03 &76.30          \\
        \xmark         & \cmark         & \cmark       & \xmark        & \xmark         & 78.95 &76.23          \\
        \xmark         & \cmark         & \cmark       & \cmark        & \xmark         & 78.48 &76.11          \\
        \xmark         & \cmark         & \cmark       & \cmark        & \cmark         & 78.41 &76.02          \\
        \xmark         & \cmark         & \xmark       & \xmark        & \cmark         & \textbf{79.13}&\textbf{76.38} \\
        \bottomrule
    \end{tabular}
    \label{tab:aug}
\end{table}

\begin{figure}[ht]
	\centering
	\begin{subfigure}{0.48\linewidth}
		\includegraphics[width=\textwidth]{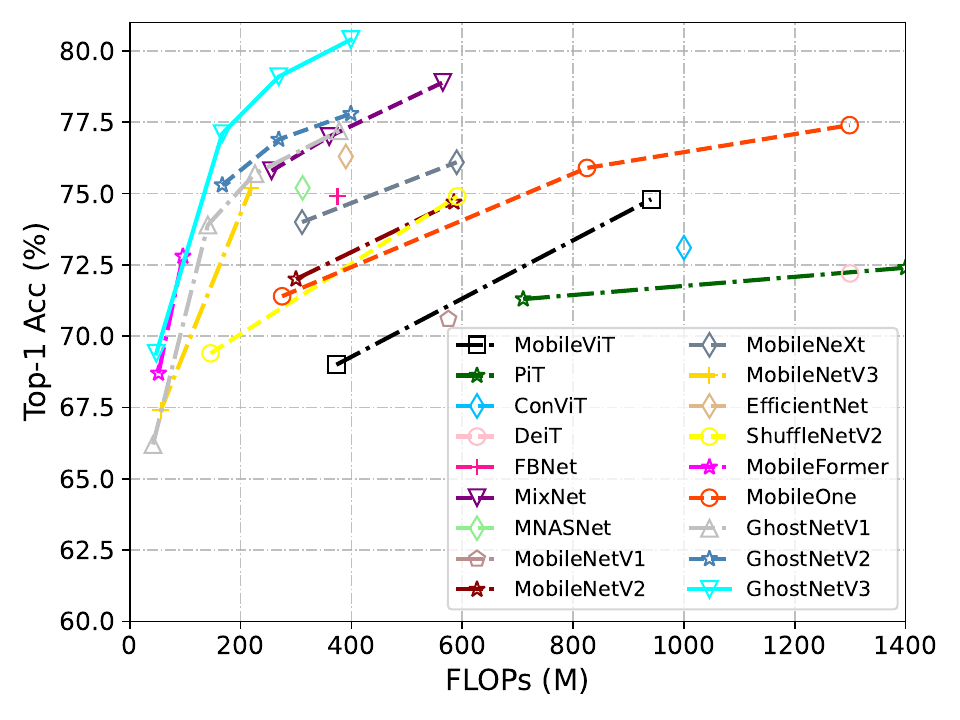}
		\caption{FLOPs}
	\end{subfigure}
	\hfill
	\begin{subfigure}{0.48\linewidth}
		\includegraphics[width=\textwidth]{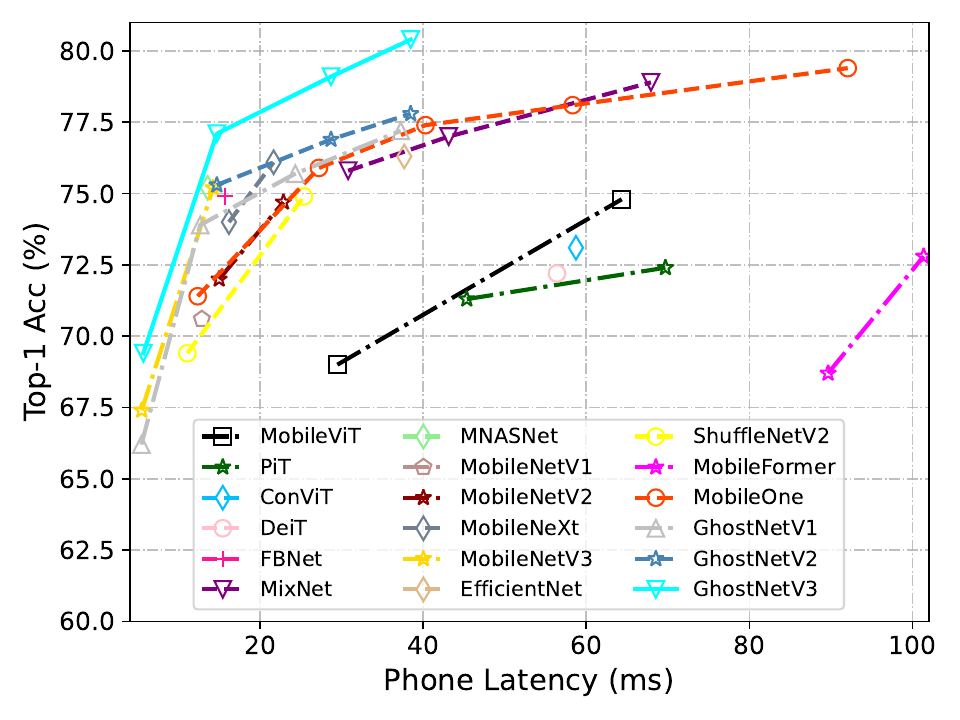}
		\caption{Mobile phone latency}
	\end{subfigure}
	\caption{The FLOPs and the latency of the compact models on mobile phone.}
	\label{fig:phone}
\end{figure}

\begin{table}[hbt!]
    \centering
    \tablestyle{3pt}{1.0}
    \caption{Top-1 accuracy achieved by various compact models on the ImageNet dataset.}
    \vspace{-0.3cm}
    \begin{tabular}{lcccccc}
        \toprule[1pt]
        \multirow{2}{*}{\textbf{Method}}             & \multirow{2}{*}{\textbf{\makecell{Params\\(M)}}} & \multirow{2}{*}{\textbf{\makecell{FLOPs\\(M)}}} & \multicolumn{2}{c}{\textbf{Latency(ms)}} & \multirow{2}{*}{\textbf{Top-1}} & \multirow{2}{*}{\textbf{Top-5}} \\
        & & & \textbf{CPU}& \textbf{Mobile} & & \\    \midrule
        GhostNetV1 0.5$\times$~\cite{ghostnet}& 2.6 & 42 & 1.98 & 5.43 & 66.2 & 86.6  \\
        Mobileformer-52~\cite{mobileformer}   & 3.6 & 52 & 34.27& 89.65& 68.7 & ---    \\
        MobileNetV3-S~\cite{mobilenetv3}      & 2.5 & 56 & 2.10 & 5.47 & 67.4 & ---    \\
        \textbf{GhostNetV3 0.5$\times$ (ours)}      & 2.7 & 48 & 2.09 & 5.67 & \textbf{69.4} & \textbf{88.5} \\
        \midrule
        
        Mobileformer-96~\cite{mobileformer}& 4.6 & 96  & 42.18 & 101.37 & 72.8 & ---  \\
        ShuffleNetV2 1.0$\times$~\cite{shufflenetv2}& 2.3 & 146 & 4.69 & 11.07 & 69.4 & 88.9  \\
        \textbf{ShuffleNetV2 1.0$\times$ + ours}    & 2.3 & 146 & 4.69 & 11.07 & 71.6 & 89.8  \\
        MobileViT-XXS~\cite{mobilevit}              & 1.3 & 373 & 12.5  & 29.47  & 69.0 & --- \\
        MobileNetV1~\cite{mobilenetv1}              & 4.2 & 575 & 7.02 & 12.85 & 70.6 & ---   \\
        MobileOne-S0~\cite{mobileone}               & 2.1 & 275 & 5.48 & 12.34 & 71.4 & ---   \\
        PiT-ti~\cite{pit}                  & 4.9 & 710 & 21.87 & 45.34  & 71.3 & ---  \\
        DeiT-tiny~\cite{deit}              & 5.9 & 1300& 28.01 & 56.4   & 72.2 & ---  \\
        PiT-xs~\cite{pit}                  & 10.6& 1400& 32.92 & 69.7   & 72.4 & ---  \\
        ConViT-tiny~\cite{convit}          & 5.7 & 1000& 28.75 & 58.73  & 73.1 & 91.7\\
        MobileViT-XS~\cite{mobilevit}      & 2.3 & 724 & 23.43 & 64.34  & 74.8 & ---  \\
        MixNet-S~\cite{mixnet}                      & 4.1 & 256 & 15.75& 30.79 & 75.8 & 92.8  \\
        MobileNetV3-L~\cite{mobilenetv3}            & 5.4 & 219 & 7.25 & 14.24 & 75.2 & ---   \\
        ShuffleNetV2 2.0$\times$~\cite{shufflenetv2}& 7.4 & 591 & 15.62& 25.36 & 74.9 & ---   \\
        MNASNet-A1~\cite{mnasnet}                   & 3.9 & 312 & 7.81 & 13.57 & 75.2 & 92.5  \\
        MobileNetV2 1.0$\times$~\cite{mobilenetv2}  & 3.4 & 300 & 7.96 & 14.97 & 72.0 & 90.8  \\
        \textbf{MobileNetV2 1.0$\times$ + ours}     & 3.4 & 300 & 7.96 & 14.97 & 75.0 & 92.1  \\
        GhostNetV1 1.0$\times$~\cite{ghostnet}      & 5.2 & 141 & 6.25 & 12.65 & 73.9 & 91.4  \\
        MobileNeXt 1.0$\times$~\cite{mobilenext}    & 3.4 & 311 & 8.63 & 16.17 & 74.0 & ---   \\
        MobileOne-S1~\cite{mobileone}               & 4.8 & 825 & 14.86& 27.21 & 75.9 & ---   \\
        FBNet-C~\cite{fbnet}                        & 5.5 & 375 & 9.37 & 15.67 & 74.9 & ---   \\
        GhostNetV2 1.0$\times$~\cite{ghostnet}      & 6.1 & 167 & 7.81 & 14.69 & 75.3 & 92.4  \\
        \textbf{GhostNetV3 1.0$\times$ (ours)}            & 6.1 & 167 & 7.81 & 14.69 & \textbf{77.1}  & \textbf{93.3} \\
        \midrule
        
        MobileNetV2 1.4$\times$~\cite{mobilenetv2} & 6.9 & 585 & 13.07 & 22.85 & 74.7 & ---   \\
        GhostNetV1 1.3$\times$~\cite{ghostnet}     & 7.3 & 226 & 12.69 & 24.31 & 75.7 & 93.4  \\
        MobileNeXt 1.4$\times$~\cite{mobilenext}   & 6.1 & 590 & 11.46 & 21.65 & 76.1 & ---   \\
        MobileOne-S2~\cite{mobileone}              & 7.8 & 1299& 23.04 & 40.28 & 77.4 & ---   \\
        MixNet-M~\cite{mixnet}                     & 5.0 & 360 & 21.44 & 43.12 & 77.0 & 93.3  \\
        MobileViT-S~\cite{mobilevit}               & 5.6 & 1792& 35.93 & 92.28  & 78.4 & ---  \\
        GhostNetV2 1.3$\times$~\cite{ghostnet}     & 8.9 & 269 & 14.46 & 28.69 & 76.9 & 93.4  \\
        \textbf{GhostNetV3 1.3$\times$ (ours)}           & 8.9 & 269 & 14.46 & 28.69 & \textbf{79.1}  & \textbf{94.5} \\ 
        \midrule
        
        GhostNetV1 1.7$\times$~\cite{ghostnet}     & 11.0 & 378 & 18.45 & 37.24 & 77.2 & 93.4 \\
        EfficientNet-B0~\cite{efficientnet}        & 5.3  & 390 & 12.50 & 37.67 & 76.3 & 93.2 \\
        GhostNetV2 1.6$\times$~\cite{ghostnetv2}   & 12.3 & 399 & 18.87 & 38.46 & 77.8 & 93.8 \\
        MobileOne-S3~\cite{mobileone}              & 10.1 & 1896& 33.31 & 58.33 & 78.1 & ---  \\
        MixNet-L~\cite{mixnet}                     & 7.3  & 565 & 34.01 & 67.93 & 78.9 & 94.2 \\
        Mobileformer-508~\cite{mobileformer}       & 14.0 & 508 & 129.58& 163.39 & 79.3 & ---  \\
        MobileOne-S4~\cite{mobileone}              & 14.8 & 2978& 52.86 & 92.06 & 79.4 & ---   \\
        \textbf{GhostNetV3 1.6$\times$ (ours)}           & 12.3 & 399 & 18.87 & 38.46 & \textbf{80.4} & \textbf{95.2} \\
        \bottomrule[1pt]
    \end{tabular}
    \label{tab:acc}
    \vspace{-0.5cm}
\end{table}

\subsection{Comparison with other compact models}

In this section, we compare GhostNetV3 with other compact models in terms of parameters, FLOPs, latency on CPU and mobile phone. Specifically, we run the models on a Windows desktop equipped with a 3.2GHz Intel i7-8700 processor to measure the CPU latency and use a Huawei Mate40Pro equipped with a Kirin 9000 CPU to evaluate the mobile phone latency under the configuration of input resolution of 224$\times$224. To ensure the lowest latency and the highest consistency, all other applications on both the CPU and mobile phone are closed. Each model is executed for 100 times to obtain reliable results.

Table~\ref{tab:acc} provides a detailed comparison of GhostNetV3 with other compact models, whose parameter counts are under 20M. From the results, the smallest transformer-based architectures require a latency of 12.5ms for inference on mobile devices, while their top-1 accuracy is only 69.0\%.
In contrast, GhostNetV3 achieves a top-1 accuracy of 77.1\% with a significantly lower latency of 7.81ms. The current state-of-the-art model, MobileFormer~\cite{mobileformer}, achieves a top-1 accuracy of 79.3\% with a latency of 129.58ms, which is unaffordable in real-world applications.
In comparison, GhostNetV3 1.6$\times$ achieves better accuracy of 80.4\% with a significantly lower latency of 18.87ms, which is 6.8$\times$ faster than MobileFormer.

Next, we compare GhostNetV3 with other CNN-based compact models, including MobileNets~\cite{mobilenetv1, mobilenetv2, mobilenetv3, mobilenext}, ShuffleNets~\cite{shufflenet, shufflenetv2}, MixNet~\cite{mixnet}, MNASNet~\cite{mnasnet}, FBNet~\cite{fbnet}, EfficientNet~\cite{efficientnet}, and MobileOne~\cite{mobileone}, where FBNet, MNASNet, and MobileNetV3 are search-based models and the others are manually designed models. Specifically, FBNet employs a hardware searching strategy while MNASNet and MobileNetV3 search the architecture parameters, such as model width, model depth, the size of the convolutional filter, \etc.

Compared with MobileNetV2~\cite{mobilenetv2}, GhostNetV3 1.0$\times$ achieves a 5.1\% improvement while maintaining almost the same latency (7.81ms vs. 7.96ms). GhostNetV3 1.3$\times$ also exhibits improved top-1 accuracy compared to MobileNeXt and EfficientNet-B0, which are 3.0\% and 2.8\%, respectively. In particular, when compared with the powerful manually designed MobileOne models, GhostNetV3 1.0$\times$ outperforms MobileOne-S1 in terms of top-1 accuracy by 1.2\% with only half the latency required. GhostNetV3 1.3$\times$ also outperforms MobileOne-S2 by 1.7\% while costing only 60\% of the latency. Moreover, when GhostNet 1.6$\times$ achieves a higher top-1 accuracy than MobileOne-S4 (80.4\% vs. 79.4\%), MobileOne's latency is 2.8$\times$ slower than that of GhostNetV3 on CPU.

When comparing GhostNetV3 1.0$\times$ with search-based compact models, it outperforms FBNet-C~cite{fbnet} by 2.2\% with faster inference speed on both CPU and phone. Additionally, GhostNetV3 1.0$\times$ provides a 1.9\% top-1 accuracy advantage over MobileNetV3 and MNASNet, while maintaining similar latency. The results demonstrate the superiority of our proposed training strategy over existing manually designed and search-based architecture designing approaches in obtaining excellent compact models.

Figure~\ref{fig:phone} presents a comprehensive performance comparison of various compact models. The left and right figures illustrate the FLOPs and the latency measured on a mobile phone, respectively. Notably, our trained GhostNetV3 stands out by exhibiting the best balance between latency and top-1 accuracy on mobile devices.

\paragraph{Other compact models}

To further demonstrate the scalability of the proposed training strategy, we apply them to the training of two other widely used compact models: ShuffleNetV2 and MobileNetV2. The results in Table~\ref{tab:acc} show that our proposed strategy can improve the top-1 accuracy of ShuffleNetV2 and MobileNetV2 by 2.2\% and 3.0\%, respectively.

\subsection{Extend to object detection}

To investigate whether the training receipts would work well for other datasets, we expand our experiments to the object detection task on COCO to validate their generalization.
The results are presented in Table~\ref{tab:od}.
Notably, the insights from the classification task are applicable to the object detection task.
For instance, the GhostNetV3 model outperforms GhostNetV2 by mAPs of 0.4 and 0.5 under the two used resolution settings, respectively.
Additionally, GhostNetV3 outperforms MobileNetV2 while requires fewer FLOPs for inference.

\begin{table}[!ht]
  \tablestyle{5pt}{1.0}
	\begin{center}
		\caption{Performance of GhostNetV3 on object detection.}
		\begin{tabular}{ccccc}
			\toprule
			Backbone               & Resolution                      & FLOPs(M) & mAP            \\ \midrule
			MobileNetV2            & \multirow{4}{*}{320$\times$230} & 613      & 22.2          \\
			GhostNetV1 1.1$\times$ &                                 & 338      & 21.8          \\
			GhostNetV2 1.0$\times$ &                                 & 342      & 22.3          \\
			GhostNetV3 1.0$\times$ &                                 & 342      & \textbf{22.7} \\ \midrule
			MobileNetV2            & \multirow{4}{*}{416$\times$416} & 1035     & 23.9          \\
			GhostNetV1 1.1$\times$ &                                 & 567      & 23.4          \\
			GhostNetV2 1.0$\times$ &                                 & 571      & 24.1          \\
			GhostNetV3 1.0$\times$ &                                 & 571      & \textbf{24.6} \\
			\bottomrule
		\end{tabular}
		\label{tab:od}
	\end{center}
  \vspace{-0.5cm}
\end{table}
\section{Conclusion}
\label{sec:con}

In this paper, we present a comprehensive study on training strategies aimed at enhancing the performance of existing compact models. The techniques, including re-parameterization, knowledge distillation, data augmentation, and learning schedule adjustments, involve no modifications to the model architecture during inference. In particular, our trained GhostNetV3 achieves an optimal balance between accuracy and inference costs, as verified on both CPU and mobile phone platforms. We also apply the proposed training strategy to other compact models such as MobileNetV2 and ShuffleNetV2, where significant improvements in accuracy are observed. We hope that our study can provide valuable insights and experiences for future research in this field.

%
%
\bibliographystyle{splncs04}
\bibliography{ref}

\begin{thebibliography}{10}
\providecommand{\url}[1]{\texttt{#1}}
\providecommand{\urlprefix}{URL }
\providecommand{\doi}[1]{https://doi.org/#1}

\bibitem{chen2021empirical}
Chen, X., Xie, S., He, K.: An empirical study of training self-supervised vision transformers. In: Proceedings of the IEEE/CVF International Conference on Computer Vision. pp. 9640--9649 (2021)

\bibitem{mobileformer}
Chen, Y., Dai, X., Chen, D., Liu, M., Dong, X., Yuan, L., Liu, Z.: Mobile-former: Bridging mobilenet and transformer. In: Proceedings of the IEEE/CVF Conference on Computer Vision and Pattern Recognition. pp. 5270--5279 (2022)

\bibitem{autoaugment}
Cubuk, E.D., Zoph, B., Mane, D., Vasudevan, V., Le, Q.V.: Autoaugment: Learning augmentation policies from data. arXiv preprint arXiv:1805.09501  (2018)

\bibitem{randaugment}
Cubuk, E.D., Zoph, B., Shlens, J., Le, Q.V.: Randaugment: Practical automated data augmentation with a reduced search space. In: Proceedings of the IEEE/CVF conference on computer vision and pattern recognition workshops. pp. 702--703 (2020)

\bibitem{imagenet}
Deng, J., Dong, W., Socher, R., Li, L., Li, K., Fei{-}Fei, L.: Imagenet: {A} large-scale hierarchical image database. In: IEEE/CVF Conference on Computer Vision and Pattern Recognition (2009)

\bibitem{ding2021diverse}
Ding, X., Zhang, X., Han, J., Ding, G.: Diverse branch block: Building a convolution as an inception-like unit. In: Proceedings of the IEEE/CVF Conference on Computer Vision and Pattern Recognition. pp. 10886--10895 (2021)

\bibitem{repvgg}
Ding, X., Zhang, X., Ma, N., Han, J., Ding, G., Sun, J.: Repvgg: Making vgg-style convnets great again. In: Proceedings of the IEEE/CVF conference on computer vision and pattern recognition. pp. 13733--13742 (2021)

\bibitem{vit}
Dosovitskiy, A., Beyer, L., Kolesnikov, A., Weissenborn, D., Zhai, X., Unterthiner, T., Dehghani, M., Minderer, M., Heigold, G., Gelly, S., et~al.: An image is worth 16x16 words: Transformers for image recognition at scale. arXiv preprint arXiv:2010.11929  (2020)

\bibitem{convit}
d’Ascoli, S., Touvron, H., Leavitt, M.L., Morcos, A.S., Biroli, G., Sagun, L.: Convit: Improving vision transformers with soft convolutional inductive biases. In: International Conference on Machine Learning. pp. 2286--2296. PMLR (2021)

\bibitem{ghostnet}
Han, K., Wang, Y., Tian, Q., Guo, J., Xu, C., Xu, C.: Ghostnet: More features from cheap operations. In: Proceedings of the IEEE/CVF conference on computer vision and pattern recognition. pp. 1580--1589 (2020)

\bibitem{resnet}
He, K., Zhang, X., Ren, S., Sun, J.: Deep residual learning for image recognition. In: Proceedings of the IEEE conference on computer vision and pattern recognition. pp. 770--778 (2016)

\bibitem{he2019bag}
He, T., Zhang, Z., Zhang, H., Zhang, Z., Xie, J., Li, M.: Bag of tricks for image classification with convolutional neural networks. In: Proceedings of the IEEE/CVF conference on computer vision and pattern recognition. pp. 558--567 (2019)

\bibitem{pit}
Heo, B., Yun, S., Han, D., Chun, S., Choe, J., Oh, S.J.: Rethinking spatial dimensions of vision transformers. In: Proceedings of the IEEE/CVF International Conference on Computer Vision. pp. 11936--11945 (2021)

\bibitem{hinton2015distilling}
Hinton, G.E., Vinyals, O., Dean, J.: Distilling the knowledge in a neural network. arXiv preprint arXiv:1503.02531  (2015)

\bibitem{mobilenetv3}
Howard, A., Sandler, M., Chu, G., Chen, L.C., Chen, B., Tan, M., Wang, W., Zhu, Y., Pang, R., Vasudevan, V., et~al.: Searching for mobilenetv3. In: Proceedings of the IEEE/CVF international conference on computer vision. pp. 1314--1324 (2019)

\bibitem{mobilenetv1}
Howard, A.G., Zhu, M., Chen, B., Kalenichenko, D., Wang, W., Weyand, T., Andreetto, M., Adam, H.: Mobilenets: Efficient convolutional neural networks for mobile vision applications. arXiv preprint arXiv:1704.04861  (2017)

\bibitem{squeezenet}
Iandola, F.N., Han, S., Moskewicz, M.W., Ashraf, K., Dally, W.J., Keutzer, K.: Squeezenet: Alexnet-level accuracy with 50x fewer parameters and< 0.5 mb model size. arXiv preprint arXiv:1602.07360  (2016)

\bibitem{shufflenetv2}
Ma, N., Zhang, X., Zheng, H.T., Sun, J.: Shufflenet v2: Practical guidelines for efficient cnn architecture design. In: Proceedings of the European conference on computer vision (ECCV). pp. 116--131 (2018)

\bibitem{mobilevit}
Mehta, S., Rastegari, M.: Mobilevit: light-weight, general-purpose, and mobile-friendly vision transformer. arXiv preprint arXiv:2110.02178  (2021)

\bibitem{peng2022beit}
Peng, Z., Dong, L., Bao, H., Ye, Q., Wei, F.: Beit v2: Masked image modeling with vector-quantized visual tokenizers. arXiv preprint arXiv:2208.06366  (2022)

\bibitem{mobilenetv2}
Sandler, M., Howard, A., Zhu, M., Zhmoginov, A., Chen, L.C.: Mobilenetv2: Inverted residuals and linear bottlenecks. In: Proceedings of the IEEE conference on computer vision and pattern recognition. pp. 4510--4520 (2018)

\bibitem{mnasnet}
Tan, M., Chen, B., Pang, R., Vasudevan, V., Sandler, M., Howard, A., Le, Q.V.: Mnasnet: Platform-aware neural architecture search for mobile. In: Proceedings of the IEEE/CVF conference on computer vision and pattern recognition. pp. 2820--2828 (2019)

\bibitem{efficientnet}
Tan, M., Le, Q.: Efficientnet: Rethinking model scaling for convolutional neural networks. In: International conference on machine learning. pp. 6105--6114. PMLR (2019)

\bibitem{tan2021efficientnetv2}
Tan, M., Le, Q.: Efficientnetv2: Smaller models and faster training. In: International conference on machine learning. pp. 10096--10106. PMLR (2021)

\bibitem{mixnet}
Tan, M., Le, Q.V.: Mixconv: Mixed depthwise convolutional kernels. arXiv preprint arXiv:1907.09595  (2019)

\bibitem{ghostnetv2}
Tang, Y., Han, K., Guo, J., Xu, C., Xu, C., Wang, Y.: Ghostnetv2: Enhance cheap operation with long-range attention. arXiv preprint arXiv:2211.12905  (2022)

\bibitem{deit}
Touvron, H., Cord, M., Douze, M., Massa, F., Sablayrolles, A., J{\'e}gou, H.: Training data-efficient image transformers \& distillation through attention. In: International conference on machine learning. pp. 10347--10357. PMLR (2021)

\bibitem{mobileone}
Vasu, P.K.A., Gabriel, J., Zhu, J., Tuzel, O., Ranjan, A.: An improved one millisecond mobile backbone. arXiv preprint arXiv:2206.04040  (2022)

\bibitem{wightman2021resnet}
Wightman, R., Touvron, H., J{\'e}gou, H.: Resnet strikes back: An improved training procedure in timm. arXiv preprint arXiv:2110.00476  (2021)

\bibitem{fbnet}
Wu, B., Dai, X., Zhang, P., Wang, Y., Sun, F., Wu, Y., Tian, Y., Vajda, P., Jia, Y., Keutzer, K.: Fbnet: Hardware-aware efficient convnet design via differentiable neural architecture search. In: Proceedings of the IEEE/CVF Conference on Computer Vision and Pattern Recognition. pp. 10734--10742 (2019)

\bibitem{lamb}
You, Y., Li, J., Reddi, S., Hseu, J., Kumar, S., Bhojanapalli, S., Song, X., Demmel, J., Keutzer, K., Hsieh, C.J.: Large batch optimization for deep learning: Training bert in 76 minutes. arXiv preprint arXiv:1904.00962  (2019)

\bibitem{cutmix}
Yun, S., Han, D., Oh, S.J., Chun, S., Choe, J., Yoo, Y.: Cutmix: Regularization strategy to train strong classifiers with localizable features. In: Proceedings of the IEEE/CVF international conference on computer vision. pp. 6023--6032 (2019)

\bibitem{mixup}
Zhang, H., Cisse, M., Dauphin, Y.N., Lopez-Paz, D.: mixup: Beyond empirical risk minimization. arXiv preprint arXiv:1710.09412  (2017)

\bibitem{shufflenet}
Zhang, X., Zhou, X., Lin, M., Sun, J.: Shufflenet: An extremely efficient convolutional neural network for mobile devices. In: Proceedings of the IEEE conference on computer vision and pattern recognition. pp. 6848--6856 (2018)

\bibitem{randomerasing}
Zhong, Z., Zheng, L., Kang, G., Li, S., Yang, Y.: Random erasing data augmentation. In: Proceedings of the AAAI conference on artificial intelligence. vol.~34, pp. 13001--13008 (2020)

\bibitem{mobilenext}
Zhou, D., Hou, Q., Chen, Y., Feng, J., Yan, S.: Rethinking bottleneck structure for efficient mobile network design. In: Computer Vision--ECCV 2020: 16th European Conference, Glasgow, UK, August 23--28, 2020, Proceedings, Part III 16. pp. 680--697. Springer (2020)

\end{thebibliography}
\end{document}